\documentclass{article}
\usepackage{ijcai16}
\usepackage{bm}
\usepackage{amsmath}
\usepackage{amssymb}
\usepackage[dvips]{graphicx}
\usepackage{algorithm}
\usepackage{algorithmic}
\usepackage{booktabs}
\usepackage{url}
\usepackage{booktabs}
\usepackage{multirow}

\usepackage{color}
\usepackage{colortbl}
\usepackage{breqn,flexisym,array}
\usepackage{mathtools}

\newtheorem{proposition}{Proposition}[section]

\newcommand{\0}{{\bm{0}}}

\newcommand{\vh}{{\bm{h}}}

\newcommand{\vu}{{\bm{u}}}

\newcommand{\vv}{{\bm{v}}}

\newcommand{\x}{{\bm{x}}}

\newcommand{\z}{{\bm{z}}}

\newcommand{\vB}{{\bm{B}}}
\newcommand{\vC}{{\bm{C}}}

\newcommand{\bE}{{\mathbb{E}}}

\newcommand{\vF}{{\bm{F}}}

\newcommand{\cH}{{\mathcal{H}}}

\newcommand{\vI}{{\bm{I}}}

\newcommand{\vM}{{\bm{M}}}

\newcommand{\cN}{{\mathcal{N}}}

\newcommand{\cQ}{{\mathcal{Q}}}
\newcommand{\vQ}{{\bm{Q}}}

\newcommand{\bR}{{\mathbb{R}}}
\newcommand{\vR}{{\bm{R}}}

\newcommand{\bS}{{\mathbb{S}}}

\newcommand{\vS}{{\bm{S}}}

\newcommand{\vU}{{\bm{U}}}

\newcommand{\vW}{{\bm{W}}}

\newcommand{\X}{{\bm{X}}}

\newcommand{\veps}{{\bm{\epsilon}}}

\newcommand{\vpsi}{{\bm{\psi}}}

\newcommand{\vLam}{{\bm{\Lambda}}}

\newcommand{\vThet}{{\bm{\Theta}}}

\DeclareMathOperator{\diag}{diag}

\DeclareMathOperator{\logdet}{logdet}

\newcommand{\argmin}{\mathop{\textrm{argmin}}\limits}

\newenvironment{tsaligned}{\begin{equation}\begin{aligned}}{\end{aligned}\end{equation}\hspace{-0.1cm}}

\newenvironment{tsaligned*}{\begin{equation*}\begin{aligned}}{\end{aligned}\end{equation*}\hspace{-0.1cm}}

\newcommand{\etal}{{et\,al.}\,}
\newcommand{\ie}{{i.\,e.},\,}
\newcommand{\eg}{{e.\,g.}\,}

\newcommand{\tslong}[1]{{#1}}
\newcommand{\tsshort}[1]{{}}

\usepackage{times}

\title{Parametric Models for Mutual Kernel Matrix Completion}
\author{Rachelle Rivero and Tsuyoshi Kato \\
Graduate School of Science and Technology, \\ 
Gunma University, \\
1--5--1 Tenjin-chou, Kiryu, Gunma 376--8515, Japan \\
rivero-rachelle@kato-lab.cs.gunma-u.ac.jp}

\begin{document}

\maketitle

\begin{abstract}
Recent studies utilize multiple kernel learning to deal with incomplete-data problem. In this study, we introduce new methods that do not only complete multiple incomplete kernel matrices simultaneously, but also allow control of the flexibility of the model by parameterizing the model matrix. By imposing restrictions on the model covariance, overfitting of the data is avoided. A limitation of kernel matrix estimations done via optimization of an objective function is that the positive definiteness of the result is not guaranteed. In view of this limitation, our proposed methods employ the LogDet divergence, which ensures the positive definiteness of the resulting inferred kernel matrix. We empirically show that our proposed restricted covariance models, employed with LogDet divergence, yield significant improvements in the generalization performance of previous completion methods.
\end{abstract}

\section{Introduction}
\label{sec:Intro}
Since the seminal work of \tsshort{Lanckriet \etal }\cite{Lanckriet2004}, data fusion has become an integral part of data analysis especially in the field of computational biology and bioinformatics. For instance, they showed that a set of proteins can be described by a number of relevant data sources, such as protein-protein interaction, gene expression, and amino acid sequences. This is because relevant data sources provide complementary perspectives or ``views'' of the objects and, together, these pieces of information present a bigger picture of the relations the objects have with each other. This notion of exploiting the multiple views of the data for better learning is more commonly known as \emph{multi-view learning}. The data sources, however, may come in various forms (\eg strings, trees, or graphs), and kernel methods~\cite{Smola,Shawe:2004} provide a way of integrating such heterogeneous data by transforming them into a common format: as kernel matrices. A Bayesian formulation for efficient multiple kernel learning was presented by \tsshort{G{\"o}nen~}\cite{Gonen}, while early works in computational biology utilized multi-view learning to classify protein functions~\cite{Deng,Lanckriet2004,LanckrietData,Noble}.

A shortcoming of multi-view learning, however, is incomplete data. Incomplete data is relatively common in almost all researches, no matter how well-designed the experiments or the data gathering methods are.  A few examples of incomplete data occurrences are: a sensor may suddenly fail and go off in a remote sensing experiment; participants may not have answered some questions in a questionnaire; and inevitable data acquisition error, among others. Analysis of incomplete data may lead to invalid conclusions, since they only give minimal insights about the objects at hand. 

Thus, in addition to dealing with the heterogeneity of the data, kernel methods are utilized in several studies to handle missing information. A data source with some missing information leads to an incomplete kernel matrix (\ie \ a matrix with missing entries); however, complete kernel matrices derived from complete data sources can be exploited to provide solutions to the incomplete-data problem. Studies addressing this problem via kernel methods have progressed over time---from completion of a kernel matrix through a single complete kernel matrix~\cite{KKT}, and through multiple complete kernel matrices~\cite{DBLP:journals/bioinformatics/KatoTA05}---to simultaneous completion of multiple incomplete kernel matrices~\cite{RivLemKat17a,Bhadra2017}. The kernel completion technique in~\cite{RivLemKat17a} associates the kernel matrices to the covariance of a zero-mean Gaussian distribution, and employs the expectation-maximization (EM) algorithm~\cite{Dempster} to minimize the objective function. On the other hand, the technique in~\cite{Bhadra2017} learns reconstruction weights to express a particular incomplete kernel matrix as a convex combination of the other kernel matrices. Although these two methods tackle a similar setting, the main difference between them is that~\cite{Bhadra2017} employ Euclidean metric to assess the distance between kernel matrices, which requires additional constraints to keep all kernel matrices positive definite. In~\cite{RivLemKat17a}, LogDet divergence~\cite{MatRelSesKat16a,DavKulJaiSraDhi07a} is employed, and this not only keeps the positive definiteness automatically but also brings a strong connection to the classical approach of estimating missing values in vectorial data. With the missing entries inferred, the completed kernel matrices can now be fused and utilized for tasks such as multi-view clustering and classification.

In the previous solution to the task of multiple kernel matrix completion~\cite{RivLemKat17a}, a model matrix is introduced as a representative kernel matrix of the given multiple kernel matrices. The model matrix is allowed to move to any point in the positive definite cone which is a very broad manifold in the set of symmetric matrices. Like the classical model fitting task, a too flexible model tends to overfit to the given empirical data. The flexibility of the model should be adjusted to make the model generalize well, but is impossible to do in the previous model. 

In view of this, we present alternative approaches to the previous method for multiple kernel matrix completion by defining parametric models that can move only on a submanifold in the positive definite cone. Both the previous and the new methods can be related to a statistical framework. The previous method can be explained with maximum likelihood estimation of a full covariance Gaussian, which often tends to overfit the data due to large degrees of freedom. On the other hand, the proposed methods can be associated to a parametric model that imposes a restriction to the covariance matrix parameter. The number of degrees of freedom in the new models can be adjusted, thereby improving the generalization performance.

\section{Problem Setting}
\label{sec:PSetting}
\begin{figure}
  \centering
  \includegraphics[width=0.3\textwidth]{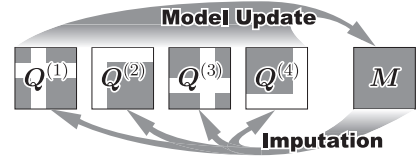}
  \caption{Overview of mutual kernel matrix completion methods. {\small In this figure, four incomplete empirical kernel matrices, $\vQ^{(1)}$, $\dots$, $\vQ^{(4)}$, are assumed to be given. A model matrix $\vM$ is introduced. The proposed method repeats two steps: model update step and imputation step. In model update step, the model matrix $\vM$ is fitted to the set of the current empirical kernel matrices $(\vQ^{(k)})_{k=1}^{4}$. In imputation step, the missing entries in each empirical kernel matrix are estimated using the current $\vM$. }
      \label{fig:probsetting}}
\end{figure}

Suppose we have $K$ data sources, some or all of which have missing information (Figure~\ref{fig:probsetting}). The algorithm works on the corresponding incomplete kernel matrices $(\bm{Q}^{(k)})_{k=1}^{K}$ of these data sources, each of size $\ell \times \ell$. The rows and columns with missing entries in each of the kernel matrices are rearranged such that the first $n_{k} < \ell$ objects in $\bm{Q}^{(k)}$ contain the available information, and unavailable for the remaining $(\ell - n_{k})$ objects. This rearrangement of the rows and columns results in the following symmetric partitioned matrix:
\begin{equation}
	\bm{Q}^{(k)}_{vh,vh} = \begin{pmatrix}
		\bm{Q}^{(k)}_{v,v}	&	\bm{Q}^{(k)}_{v,h} \cr
		\bm{Q}^{(k)}_{h,v}	&	\bm{Q}^{(k)}_{h,h} \cr
		\end{pmatrix},\label{eq:Q}
\end{equation}
where $\bm{Q}^{(k)}_{v,v} \in \mathbb{S}^{n_{k}}_{++}$, with $\mathbb{S}^{n_{k}}_{++}$ denoting the set of $n_{k} \times n_{k}$ strictly positive definite symmetric matrices. The algorithm then mutually infers the (missing) entries for the submatrices $\bm{Q}^{(k)}_{v,h} \in \mathbb{R}^{n_{k} \times \left(\ell - n_{k}\right)}$, $\bm{Q}^{(k)}_{h,v} = \left(\bm{Q}^{(k)}_{v,h}\right)^{\top}$, and $\bm{Q}^{(k)}_{h,h}\in \mathbb{S}^{\left(\ell - n_{k}\right)}_{++}$, for $k = 1,\ldots,K$.

\section{FC-MKMC: Existing Model}
\label{sec:MKMC}
In the previous study~\cite{RivLemKat17a}, an algorithm for mutual kernel matrix completion had already been developed. Henceforth, we will refer to the previous method as the full covariance mutual kernel matrix completion (FC-MKMC), and review the method in this section. To infer the missing values in incomplete kernel matrices, FC-MKMC introduces an $\ell\times \ell$ model matrix $\vM$, and finds the set of kernel matrices $\vQ^{(k)}$ that are as close to each other as possible through the model matrix $\vM$. The objective function of FC-MKMC is the sum of LogDet divergences~\cite{MatRelSesKat16a,DavKulJaiSraDhi07a}:
\begin{tsaligned}
  J_{\text{FC}}\left(\cH,\vM\right)
  \coloneqq
  \sum_{k=1}^{K}\text{LogDet}\left(\vQ^{(k)},\vM\right),
\end{tsaligned}
where $\cH\coloneqq\left\{ \vQ^{(k)}_{v,h}, \vQ^{(k)}_{h,h} \right\}_{k=1}^{K}$
is the set of submatrices containing missing entries, and
$\vM$ is the model matrix. The LogDet divergence is defined as
%
%
\begin{tsaligned}
  \text{LogDet}\left(\vQ^{(k)},\vM\right)
  &\coloneqq
  \dfrac{1}{2}\left(\logdet\vM - \logdet\vQ^{(k)}\right.
  \\
  & \qquad \left.+ \left\langle \vM^{-1},\vQ^{(k)}-\vM\right\rangle \right). 
\end{tsaligned}
An advantage of using LogDet divergence is that a necessary property for valid kernel matrices, the positive definiteness, is ensured for the resultant completed kernel matrices. 
The approach of FC-MKMC is essentially similar to the well-known probabilistic approach for classical incomplete data completion~\cite{McLachlan08book}, where missing values in incomplete vectorial data are to be inferred. In the approach for the classical task, a probabilistic model is introduced to be fitted to the temporarily completed data, and the missing values are imputed with the most probable values using the current inference of the probabilistic model. The number of degrees of freedom of a probabilistic model provides an important perspective for the success or for the failure of data completion: too rigid models cannot capture the underlying data distribution, while too flexible models are often overfitted to the data set. 
In FC-MKMC, the model matrix can take any values without restriction, with $(\ell+1)\ell/2$ degrees of freedom. This model may be too flexible. In the next section, we shall present two new models in which the number of degrees of freedom can be adjusted.

\section{Parametric Models}
The model matrix of FC-MKMC is too flexible and is not tunable.
Hence, we introduce two types of the model matrix: PCA model and FA model.
\subsection{PCA-MKMC}
\label{sec:PPCA}
In the PCA model, the form of the model matrix is restricted to
\begin{tsaligned}
  \vM:=\vW\vW^\top+\sigma^{2}\vI,
\end{tsaligned}
where the matrix $\vW\in\bR^{\ell\times q}$ and scalar $\sigma^{2}\in\bR$ are the adaptive parameters of the PCA model. The number of columns in $\vW$, say $q$, is arbitrary. Larger $q$ yields a more flexible model and vice-versa. The number of degrees of freedom of this model is $\ell q + 1 - (q-1)q/2$. 

Meanwhile, the objective function of PCA model is expressed as
\begin{tsaligned}
  J_{\text{PCA}}(\cH,\vW,\sigma^{2})
  \coloneqq
  \sum_{k=1}^{K}\text{LogDet}(\vQ^{(k)},\vW\vW^\top+\sigma^{2}\vI). 
\end{tsaligned}
Since the objective function is not jointly convex of the three arguments, $\cH$, $\vW$, and $\sigma^{2}$, the optimal solution cannot be given in closed form.  Hence, we adopt the following block coordinate descent method that repeats
the following two steps:
\begin{enumerate}
\item Imputation step: 
  \begin{tsaligned}
    \cH^{(t)} \coloneqq \argmin_{\cH}J_{\text{PCA}}(\cH,\vW^{(t-1)},\sigma^{2}_{t-1});
  \end{tsaligned}
\item Model update step: 
  \begin{tsaligned}
    (\vW^{(t)},\sigma^{2}_{t})
    \coloneqq \argmin_{(\vW,\sigma^{2})}J_{\text{PCA}}(\cH^{(t)},\vW,\sigma^{2}).
  \end{tsaligned}
\end{enumerate}
Therein, the iteration number $t$ is the superscript of $\cH$ and $\vW$,
and the subscript of $\sigma^{2}$. 
By letting $\vM\coloneqq\vW\vW^\top+\sigma^{2}\vI$, the imputation
step can be
performed in the same fashion as that of
FC-MKMC~\cite{RivLemKat17a}. 
For each data source, the rows and the columns in $\vM$
are reordered as $\vQ^{(k)}_{vh,vh}$, and partitioned
to obtain $\vM^{(k)}_{v,v}$, $\vM^{(k)}_{v,h}$, and
$\vM^{(k)}_{h,h}$.  Using these submatrices
in the model matrix, and the known submatrix in
the empirical matrix $\vQ^{(k)}_{v,v}$,
the unknown submatrices in $\vQ^{(k)}_{vh,vh}$
are re-estimated as
\begin{tsaligned}\label{eq:upd-Qvh}
  \vQ_{v,h}^{(k)} \coloneqq \vQ_{v,v}^{(k)}\left(\vM_{v,v}^{(k)}\right)^{-1}\vM_{v,h}^{(k)}
\end{tsaligned}
and
\begin{multline}\label{eq:upd-Qhh}
  \vQ_{h,h}^{(k)} \coloneqq \vM_{h,h}^{(k)} - \vM_{h,v}^{(k)}\left(\vM_{v,v}^{(k)}\right)^{-1}\vM_{v,h}^{(k)}+
  \\\vM_{h,v}^{(k)}\left(\vM_{v,v}^{(k)}\right)^{-1}\vQ_{v,v}^{(k)}\left(\vM_{v,v}^{(k)}\right)^{-1}\vM_{v,h}^{(k)}. 
\end{multline}
Finally, the submatrices are reordered back to $\vQ^{(k)}$ to obtain a new solution
that minimizes $J_{\text{PCA}}(\cH,\vW,\sigma^{2})$ over
missing values $\cH$,
with the model parameters $\vW$ and $\sigma^{2}$ held fixed.
Denote by $\vQ^{(t,k)}$ the new value of $\vQ^{(k)}$ at $t$-th iteration. 

In the model update step, the $K$ empirical kernel matrices are fixed, and the two model parameters, $\vW$ and $\sigma^{2}$, are optimized. We here denote by `const' the terms independent of $(\vW, \sigma^{2})$, and so the objective function can be rewritten as
\begin{multline}
  J_{\text{PCA}}(\cH^{(t)},\vW,\sigma^{2})
  \coloneqq
  \frac{K}{2}\logdet(\vW\vW^\top+\sigma^{2}\vI)
  \\
  +
  \frac{K}{2}
  \left<
  (\vW\vW^\top+\sigma^{2}\vI)^{-1},
  \vS^{(t)}
  \right>
  +
  \text{const}
\end{multline}
where we have defined 
\begin{tsaligned}
  \vS^{(t)} \coloneqq \frac{1}{K}\sum_{k=1}^{K}\vQ^{(t,k)}. 
\end{tsaligned}
Even though the missing values $\cH$ are fixed to $\cH^{(t)}$, the function $(\vW,\sigma^{2})\mapsto J_{\text{PCA}}(\cH^{(t)},\vW,\sigma^{2})$ is still not convex on the space of the model parameters $(\vW,\sigma^{2})$. Nevertheless, surprisingly enough, the joint optimal solution of the two model parameters $\vW$ and $\sigma^{2}$ is given in closed form~\cite{TipBis99}. Let $\lambda_{1}$, $\dots$, $\lambda_{\ell}$ be the eigenvalues of $\vS^{(t)}$. Assume that $\lambda_{1}\ge\dots\ge\lambda_{\ell}$, and denote by $\vu_{1},\dots,\vu_{\ell}$ their corresponding eigenvectors. It can be shown that the optimal $\sigma^{2}$, denoted by $\sigma^{2}_{t}$, is expressed as
\begin{tsaligned}\label{eq:upd-powsig}
  \sigma^{2}_{t} = \frac{1}{\ell-q}\sum_{j=q+1}^{\ell}\lambda_{j}. 
\end{tsaligned}
Now, let $\vU_{q}\coloneqq\left[\vu_{1},\dots,\vu_{q}\right]$ and
$\vLam_{q}\coloneqq\diag(\lambda_{1},\dots,\lambda_{q})$.
Here, for vector $\x\in\bR^{n}$, we denote the diagonal matrix with diagonal entries $\x$ by $\diag(\x)$.
Meanwhile, $\diag(\X)$ denotes an $n$-dimensional vector containing the diagonal entries in a square matrix $\X\in\bR^{n\times n}$.
And so, the optimal value of $\vW$, denoted by $\vW^{(t)}$, is given by
\begin{tsaligned}\label{eq:upd-W}
  \vW^{(t)} =
  \vU_{q} (\vLam_{q}-\sigma^{2}_{t}\vI)^{1/2}\vR,
\end{tsaligned}
where $\vR$ is an arbitrary orthonormal matrix
(\ie $\vR^\top\vR=\vI$). 

\begin{algorithm}[th!]
\caption{
PCA-MKMC Algorithm.
\label{alg:KMC}}
\begin{algorithmic}[1]
  \REQUIRE
  Kernel matrices $\left(\vQ^{(k)}\right)_{k = 1}^{K}$. 
  \ENSURE Completed kernel matrices $\left(\vQ^{(k)}\right)_{k = 1}^{K}$.
  \STATE \textbf{begin}
  \STATE Initialize $\left(\vQ^{(k)}\right)_{k = 1}^{K}$ by imputing zeros in the missing entries;
  \STATE Initialize the model matrix as $\vM=\sum_{k=1}^{K} \vQ^{(k)}/K$; 
  \REPEAT
  \FORALL{$k \in \{1,\dots,K\}$}
\STATE Apply \eqref{eq:upd-Qvh} and \eqref{eq:upd-Qhh} to update $\vQ_{v,h}^{(k)}$ and $\vQ_{h,h}^{(k)}$; 
  \ENDFOR
  \STATE Apply \eqref{eq:upd-powsig} and \eqref{eq:upd-W} to update
  $\vW$ and $\sigma^{2}$;
  \STATE $\vM:=\vW\vW^\top+\sigma^{2}\vI$; 
  \UNTIL{convergence}
  \STATE \textbf{end.}
\end{algorithmic}
\end{algorithm}

\begin{algorithm}[th]
\caption{
FA-MKMC Algorithm.
\label{alg:FA-MKMC}}
\begin{algorithmic}[1]
  \STATE As in Algorithm~\ref{alg:KMC}, except replacing line 8 by:
  Apply \eqref{eq:W-psi-upd-famkmc}
  to update $\vW$ and $\vpsi$; and line 9 by \eqref{eq:M-FA}.
\end{algorithmic}
\end{algorithm}

\subsection{FA-MKMC}
In this section, we introduce FA model, which is an alternative variant of PCA model. 
FA model uses the following parametric model as a model matrix:
\begin{tsaligned}\label{eq:M-FA}
  \vM = \vW\vW^\top + \diag(\vpsi). 
\end{tsaligned}
The difference of this from the PCA model is the second term.
In PCA model, the second term is $\sigma^{2}\vI$, whereas
in FA model, the term can take any diagonal matrix.
The number of degrees of freedom of FA model is
$\ell q + \ell - (q-1)q/2$, and 
the objective function is expressed as
\begin{tsaligned}
  J_{\text{FA}}(\cH,\vW,\vpsi)
  \coloneqq
  \sum_{k=1}^{K}\text{LogDet}\left(\vQ^{(k)},\vW\vW^\top+\diag(\vpsi)\right). 
\end{tsaligned}
Similar to the fitting algorithm of PCA model,
we adopt the block coordinate descent method to fit FA model
to empirical kernel matrices. The imputation step is
same as PCA model.  In PCA model, when fixing $\cH$,
the optimal $(\vW,\sigma^{2})$ can be expressed
in closed form.  However, in FA model,
the optimal $(\vW,\vpsi)$ cannot be given
in closed form even if $\cH$ is fixed.
In the FA model, we just improve
$(\vW,\vpsi)$ at the model update step.

The two steps in $t$-th iteration are summarized as follows.
\begin{enumerate}
\item Imputation step:
  Use \eqref{eq:upd-Qvh} and \eqref{eq:upd-Qhh} 
  to infer the missing entries $\cH^{(t)}$ such that 
  \begin{tsaligned}
    \cH^{(t)} \coloneqq \argmin_{\cH}J_{\text{FA}}(\cH,\vW^{(t-1)},\vpsi^{(t-1)});
  \end{tsaligned}
\item Model update step:
  Update model parameters $(\vW^{(t)},\vpsi^{(t)})$
  such that 
  \begin{tsaligned}\label{eq:ineq-fa-upd}
	\lefteqn{
    J_{\text{FA}}(\cH^{(t)},\vW^{(t)},\vpsi^{(t)})
	}\quad  \\	
		&\le J_{\text{FA}}(\cH^{(t)},\vW^{(t-1)},\vpsi^{(t-1)}).
  \end{tsaligned}
\end{enumerate}
A new value $(\vW^{(t)},\vpsi^{(t)})$ that satisfies
\eqref{eq:ineq-fa-upd}
can be found as follows:
Update the factor loading matrix~$\vW^{(t)}$ and the
noise variance vector $\vpsi^{(t)}$, by
\begin{tsaligned}\label{eq:W-psi-upd-famkmc}
  &\vW^{(t)} \coloneqq \vS_{xz}^{(t)}\left(\vS_{zz}^{(t)}\right)^{-1}, \qquad \text{and}
  \\
  &\vpsi^{(t)}
  \coloneqq
  \diag
  \left(
  \vS_{xx}^{(t)} - \vS_{xz}^{(t)}(\vS_{zz}^{(t)})^{-1}(\vS_{xz}^{(t)})^\top
  \right),
\end{tsaligned}
respectively, where
\begin{tsaligned}
  &\vF^{(t)} \coloneqq\left(\vW^{(t-1)}\right)^\top\diag\left(\vpsi^{(t-1)}\right)^{-1},
  \\
  &\vC^{(t)} \coloneqq \vI + \vF^{(t)}\vW^{(t-1)}, 
  \\
  &(\vM^{(t)})^{-1}
  \coloneqq
  \diag(\vpsi^{(t-1)})^{-1} - \left(\vF^{(t)}\right)^\top\left(\vC^{(t)}\right)^{-1}\vF^{(t)},
  \\
  &\vB^{(t)} \coloneqq \left(\vW^{(t-1)}\right)^\top\left(\vM^{(t)}\right)^{-1},
  \\
  &\vS_{xz}^{(t)} \coloneqq \vS^{(t)}\left(\vB^{(t)}\right)^\top,
  \\
  &\vS_{zz}^{(t)} \coloneqq \vI - \vB^{(t)}\vW^{(t-1)} + 
  \vB^{(t)}\vS_{xz}^{(t)}. 
\end{tsaligned}

\begin{proposition}\label{prop:FAMKMC-monodecreasing}
  The inequality~\eqref{eq:ineq-fa-upd} always holds
  at every iteration in Algorithm~\ref{alg:FA-MKMC}. 
\end{proposition}
\tslong{(See Sect.~\ref{ss:proof-prop:FAMKMC-monodecreasing} for
proof of Proposition~\ref{prop:FAMKMC-monodecreasing}.) } 
This proposition guarantees the monotonic decrease
of the objective value $J_{\text{FA}}(\cH^{(t)},\vW^{(t)},\vpsi^{(t)})$
during optimization. 

\section{Statistical Interpretation}
\label{sec:stat}
As described in~\cite{RivLemKat17a}, FC-MKMC falls in a statistical framework. Concretely, FC-MKMC is an algorithm that performs the maximum likelihood estimation of a model parameter $\vM$ of a probabilistic model $p_{\text{FC}}(\x\,|\,\vM) \coloneqq \cN(\x\,;\,\0,\vM)$, where $\x$ is an $\ell$-dimensional random variate.  In the statistical framework for FC-MKMC, maximum likelihood estimation is performed by finding the maximizer of the log-likelihood function
\begin{tsaligned}\label{eq:llfun-fc-def}
  L_{\text{FC}}(\vM)
  \coloneqq
  \sum_{k=1}^{K}\bE_{q_{k}(\vv_{k})}
  \left[\log \cN\left(\vv_{k}\,;\,\0,\vM^{(k)}_{v,v}\right) \right]
\end{tsaligned}
over the model parameter $\vM^{(k)}_{v,v}\in\bS^{n_{k}}_{++}$.
Therein,
$\vv_{k}\in\bR^{n_{k}}$ is the sub-vectorial variate in $\x_{k}$
associated with the visible objects in $k$-th data source;
$\vM^{(k)}_{v,v}$ is the submatrix of $\vM$
associated with $\vv_{k}$; and
$q_{k}(\cdot)$ is the empirical distribution associated
with the $k$-th data source such that the second moments 
satisfy 
\begin{tsaligned}
  \bE_{q_{k}(\vv_{k})}
  \left[
    \vv_{k}\vv_{k}^\top
    \right]
  =
  \vQ^{(k)}_{v,v}. 
\end{tsaligned}
From the log-likelihood function defined in
\eqref{eq:llfun-fc-def}, 
it is possible to derive an EM algorithm in which
E-step computes the expected value
$\bE\left[\x_{k}\x_{k}^\top\right]$
by \eqref{eq:upd-Qvh} and \eqref{eq:upd-Qhh},
based on the current model parameter.
In the EM algorithm, 
M-step updates the model parameter $\vM$ by
the maximizer of the expected complete-data log-likelihood
function~\cite{McLachlan08book} over $\vM$.
However, the model matrix is too flexible, and therefore
may be overfitted to the given empirical data. 

\subsection{EM Algorithm for PCA Model}
Here, we present a connection between FC-MKMC algorithm and the classical statistical approach for missing value estimation.
Let us discuss the case of replacing the full covariance model with the probabilistic principal component analysis (PPCA) model introduced in~\cite{TipBis99}.
When employing PPCA model with the mean parameter fixed to zero, 
the probabilistic densities of the $n_{k}$-dimensional
random variate $\vv_{k}$ are defined as 
\begin{tsaligned}
  &p_{\text{PCA}}\left(\vv_{k}\,|\,\vW,\sigma^{2}\right)
  \\
  &\coloneqq
  \int \cN\left( \x_{k}\,;\,\0, \vW\vW^\top+\sigma^{2}\vI_{\ell}\right)
  d\vh_{k}
  \\
  &=\cN\left( \vv_{k}\,;\,\0,
  \vW_{v}^{(k)}\left(\vW_{v}^{(k)}\right)^\top+\sigma^{2}\vI_{n_{k}}\right), 
\end{tsaligned}
where $\vW_{v}^{(k)}\in\bR^{n_{k}\times q}$ is the submatrix of $\vW$
containing the rows associated with the visible objects. 
The log-likelihood function of this model is given by
\begin{tsaligned}\label{eq:llfun-pca-def}
  L_{\text{PCA}}\left(\vW,\sigma^{2}\right) &\coloneqq \sum_{k=1}^{K}\bE_{q_{k}(\vv_{k})}
  \left[\log p_{\text{PCA}}\left(\vv_{k}\,|\,\vW,\sigma^{2}\right)\right],
\end{tsaligned}
which is used in finding the maximum likelihood estimate (MLE) of the model parameters $\vW$ and $\sigma^{2}$ of the PPCA model.
The expected complete-data log-likelihood function,
also known as the Q-function, 
can be written as
\begin{tsaligned}
  &\cQ^{\text{PCA}}_{t}(\vW,\sigma^{2})
  \coloneqq
  \sum_{k=1}^{K}\bE\left[\log p_{\text{PCA}}(\x_{k}\,|\,\vW,\sigma^{2}\vI)\right]
  \\
  &=
  -
  \frac{K}{2}\logdet\left(\vW\vW^\top+\sigma^{2}\vI\right)
  \\
  &\quad -\frac{1}{2}
  \left<
  \left(\vW\vW^\top+\sigma^{2}\vI\right)^{-1},
  \sum_{k=1}^{K}
  \bE\left[\x_{k}\x_{k}^\top\right]
  \right>,
\end{tsaligned}
where we have dropped the terms that do not depend on
the model parameters.  Therein, the operator $\bE$
takes mathematical expectation 
under the joint posterior densities $q(\vh_{k}|\vv_{k})q(\vv_{k})$
defined from the current value of $\vW$ and $\sigma^{2}$. 
By letting $\vQ^{(k)} \coloneqq \bE\left[\x_{k}\x_{k}^\top\right]$,
the negative Q-function is equal to $J_{\text{PCA}}$
up to constants, implying that the M-step of the EM algorithm
is given by \eqref{eq:upd-powsig} and \eqref{eq:upd-W}.  
Hence, we can say that the PCA-MKMC
algorithm presented in the previous section is an EM algorithm.

\subsection{EM Algorithm for FA Model}
This section is concluded by showing that FA-MKMC is an EM algorithm for fitting the probabilistic factor analysis (PFA) model~\cite{bartholomew2008analysis}.
In the PFA model, a latent variable vector $\z_{k}\in\bR^{q}$, drawn from the isotropic Gaussian $\cN(\0,\vI_{q})$, is introduced for each data source. Then, $\x_{k}$ is generated by the process $\x_{k}=\vW\z_{k}+\veps_{k}$, where $\veps_{k}$ is a Gaussian noise drawn from $\cN(\0,\text{diag}(\vpsi))$. 
For this FA model, we treat $(\x_{k},\z_{k})$ as the complete data for $k$-th data source to develop an EM algorithm for maximum likelihood estimation. 
The probabilistic densities of the $n_{k}$-dimensional random variate $\vv_{k}$ are obtained by marginalizing $\vh_{k}$ and $\z_{k}$ out from the joint densities of the complete data:  
\begin{multline}
  p_{\text{FA}}(\x_{k},\z_{k}\,|\,\vW,\vpsi)
  \\
  \coloneqq
  \cN\left( \x_{k}\,;\,\vW\z_{k},\diag(\vpsi)\right)
  \cN\left( \z_{k}\,;\,\0,\vI_{q} \right). 
\end{multline}
The Q-function is written as
\begin{tsaligned}\label{eq:Qfun-fa}
  \cQ^{\text{FA}}_{t}(\vW,\vpsi)
  &\coloneqq
  \sum_{k=1}^{K}\bE\left[\log p_{\text{FA}}(\x_{k},\z_{k}\,|\,\vW,\vpsi)\right]
  \\
  &=
  - \left<\sum_{k=1}^{K}\bE\left[\z_{k}\x_{k}^\top\right],
    \diag(\vpsi)^{-1} \vW
    \right>
  \\
  &\quad
  -\frac{1}{2}\left<
  \sum_{k=1}^{K}\bE\left[\z_{k}\z_{k}^\top\right],
  \vW^\top \diag(\vpsi)^{-1} \vW \right>
    \\
    &\quad-\frac{1}{2}\left< 
  \sum_{k=1}^{K}\bE\left[\x_{k}\x_{k}^\top\right],
  \diag(\vpsi)^{-1}\right>
  \\
  &\quad-\frac{K\ell}{2}\log(2\pi)
  -\frac{K}{2}\sum_{i=1}^{\ell}\log\psi_{i} ,
\end{tsaligned}
where $\bE$ here is 
the mathematical expectation that operates 
under the joint posterior densities $q_{t}(\z_{k},\vh_{k}|\vv_{k})q(\vv_{k})$
depending on the current value of $\vW$ and $\vpsi$ obtained
at the $(t-1)$ iteration.
It can be shown that, by letting
$\vQ^{(k)} = \bE\left[\x_{k}\x_{k}^\top\right]$, 
the expected values computed in the $t$-th iteration
in the EM algorithm are expressed as
\begin{tsaligned}\label{eq:faem-expected-is-S}
  \sum_{k=1}^{K}\bE\left[\z_{k}\z_{k}^\top\right] = K\vS^{(t)}_{zz} \quad
  \text{and}\quad
  \sum_{k=1}^{K}\bE\left[\z_{k}\x_{k}^\top\right] = K(\vS^{(t)}_{xz})^\top. 
\end{tsaligned}
In the M-step of EM algorithm, the model parameters $(\vW,\vpsi)$ that maximizes the Q-function are found. Setting the derivative of the Q-function with respect to the model parameters to zero, it turns out that the optimal factor loading matrix $\vW$ and noise variance vector $\vpsi$ are given by \eqref{eq:W-psi-upd-famkmc}. Hence, FA-MKMC is an EM algorithm.
\tslong{(See Sect.~\ref{ss:estep-famkmc} and Sect.~\ref{ss:mstep-famkmc} for
derivations of E-step and M-step, respectively. )}

\begin{table*}[t!]
  \caption{\small Classification performance of the completion methods for 20\% missed kernel data. The table entries are the ROC scores for a given functional class, averaged over ten trials. Here, the SVM classifier is trained on 20\% of the combined completed kernel matrices. The boldfaced values correspond to the largest ROC score in each row, while the underlined values correspond to the ROC scores with no significant difference from the highest ROC score in each class.}
  \label{tab:ROC}
  \begin{center}
    \scalebox{0.85}{
    \begin{tabular}{cccccccc}
      %
      \toprule
    Class &	zero-SVM						&	mean-SVM						&	FC-MKMC					&	PCA-GK							&	PCA-K								& FA-GK								&	FA-K								\\\hline	
				1	&	0.7914							&	0.7915							&	0.7995					&	0.8015							&	\textbf{0.8022}			&	0.8010							&	0.8006							\\
				2	&	0.7918							&	0.7925							&	0.7975					&	\underline{0.8025}	&	\textbf{0.8032}			&	0.8014							&	\underline{0.8021}	\\
				3	&	0.7941							&	0.7933							&	0.8000					&	\underline{0.8045}	&	\textbf{0.8052}			&	0.8029							&	0.8032							\\
				4	&	0.8418							&	0.8431							&	0.8497					&	0.8529							&	\textbf{0.8534}			&	0.8516							&	0.8519							\\
				5	&	0.8839							&	0.8844							&	0.8956					&	0.8972							&	\textbf{0.8979}			&	0.8961							&	0.8967							\\
				6	&	0.7665							&	0.7669							&	0.7745					&	\underline{0.7780}	&	\textbf{0.7783}			&	\underline{0.7770}	&	0.7770							\\
				7	&	0.8321							&	0.8328							&	0.8414					&	0.8437							&	\textbf{0.8444}			&	0.8429							&	\underline{0.8440}	\\
				8	&	0.7336							&	0.7336							&	0.7354					&	\underline{0.7407}	&	\textbf{0.7418}			&	0.7391							&	0.7386							\\
				9	&	0.7621							&	0.7630							&	0.7651					&	\underline{0.7706}	&	\textbf{0.7714}			&	\underline{0.7694}	&	\underline{0.7695}	\\
			10	&	0.7441							&	0.7445							&	0.7485					&	0.7551							&	\textbf{0.7570}			&	0.7525							&	\underline{0.7556}	\\
			11	&	0.5766							&	0.5757							&	\textbf{0.5825}	&	\underline{0.5791}	&	\underline{0.5807}	&	0.5793							&	0.5772							\\
			12	&	0.9357							&	0.9347							&	0.9435					&	\underline{0.9448}	&	\textbf{0.9453}			&	0.9443							&	0.9444							\\
			13	&	\underline{0.6818}	&	\underline{0.6845}	&	0.6794					&	\textbf{0.6913}			&	\underline{0.6911}	&	0.6840							&	0.6838							\\\bottomrule
    \end {tabular}
    }
  \end{center}
\end{table*}

\section{Experimental Settings}
\label{sec:Exp}
To test how much information the kernel matrices will retain after the completion processes, we subject the completed kernel matrices to a classification task: the functional classification prediction of yeast proteins. For this task, a collection of six kernel matrices representing different data types is used:
the enriched kernel matrix $\bm{K}_{\text{Pfam'}}$; the three interaction kernel matrices $\bm{K}_{\text{Gen}}$, $\bm{K}_{\text{Phys}}$, and $\bm{K}_{\text{TAP}}$; a Gaussian kernel defined directly on gene expression profiles $\bm{K}_{\text{Exp'}}$; and the Smith-Waterman matrix $\bm{K}_{\text{SW}}$; as described in \cite{LanckrietData}. While each kernel representation contains partial information on the similarities among yeast proteins, the combination of these kernel matrices is known to provide a bigger picture of the relationships among these proteins through the different views of the data \cite{Lanckriet2004,LanckrietData,Noble}---making the combined form more suitable in the overall predictions for the function classification task.  

Meanwhile, the 13 functional classes considered are listed in \cite{LanckrietData}, which include metabolism, transcription, and protein synthesis, among others. If, for example, a certain protein is known to carry out metabolism and protein synthesis, then this protein is labeled as $+1$ in these categories and $-1$ elsewhere. This setting can then be viewed as 13 binary classification tasks.

In this study we utilized $K=6$ related data sources for function prediction of $\ell=3,588$ yeast proteins. Initially, the kernel matrices may have missing rows and columns, which correspond to some missing information about the relationships among yeast proteins in the data sources. Our goal is to infer the missing entries in the kernel matrices, whilst retaining as much valuable information about the protein relationships as possible. 

Our experiments consist of two stages: the kernel matrix completion (or missing data inference) stage, and the classification stage---the details of which are given in the subsequent sections.

\subsection{Data Inference Stage}\label{sec:stage1}
In this stage, mutual completion of the kernel matrices is performed. Since our data set has no missing entries, we generated incomplete kernel matrices by artificially removing some entries, following the process in \cite{RivLemKat17a}. Here, rows and (corresponding) columns were randomly picked, and undetermined values (zeros for zero-imputation method, and unconditional mean for mean-imputation method) were imputed; the details of which are referred to in \cite{RivLemKat17a}. For numerical stability of the two EM-based methods, $\vS$ is transformed to $(K\vS+10^{-3}\bm{I})/(K+10^{-3})$ at each iteration, a trick that is often used in Gaussian fitting. In our experiments, different percentages of missing entries were considered, and the incomplete kernel matrices were initialized by zero-imputation before proceeding with the completion processes, as specified in Alg.~\ref{alg:KMC}.

\subsection{Classification Stage}\label{sec:stage2}
After the completion process, a support vector machine (SVM)~\cite{cristianini2000,Smola} is used to predict whether a yeast protein belongs to a certain functional class or not. Since a yeast protein is not limited to a single functional class, the prediction problem is structured as 13 binary classification tasks, where an SVM classifier is trained on 20\% randomly-picked data points on the combined kernel matrices. We then assess, in each functional class, the classification performance of the algorithms via receiver operator characteristic (ROC)---a widely-used performance measure for imbalanced data sets. Higher ROC score means better classification performance. The experiments were performed ten times, and the averages of ROC scores across the ten trials are recorded in Tab.~\ref{tab:ROC}.

\section{Experimental Results}
\label{sec:Res}
In this section we present experimental comparisons among the five multiple kernel completion techniques: zero-SVM, mean-SVM, FC-MKMC, PCA-MKMC, and FA-MKMC. We refer to the completion methods zero-imputation and mean-imputation as zero-SVM and mean-SVM, respectively, after an SVM classifier has been trained. In our experiments, we used two criteria in choosing the number $q$ of principal components for PCA and FA models: the Guttman-Kaiser and Kaiser criterion, where $q$ is the number of all eigenvalues greater than the mean of the eigenvalues and greater than one, respectively~\cite{Jolliffe:1986}. Henceforth, we will use PCA-GK and PCA-K to refer to PCA-MKMC, and FA-GK and FA-K to refer to FA-MKMC, with principal components via Guttman-Kaiser criterion and Kaiser criterion, respectively.

The ROC scores of the completion methods after inferring the missed 20\% of the data are summarized in Tab.~\ref{tab:ROC}. Note, however, that the experiments were done on different percentages of missed entries; but due to lack of space, the results on the other cases are reported in a longer version of this paper.

In the case of completion of 20\% missed data, the proposed method of restricting the model covariance achieves the highest ROC score in all classes, except at the 11th functional class where FC-MKMC obtains the highest ROC score; and in this case PCA-GK and PCA-K has no statistical difference from FC-MKMC according to a one-sample $t$-test. It can also be noted that in most cases, the classification performance of the restricted covariance models are not significantly lower than the highest ROC scores.


\section{Conclusion}
\label{sec:Concl}
In this study we present new methods, called PCA-MKMC and FA-MKMC, to solve the problem of mutually inferring the missing entries of kernel matrices, while controlling the flexibility of the model. In contrast to the full-covariance model parameter in the existing method, our algorithm imposes a restriction to the model covariance, capturing only the most relevant information in the data set through the principal components or factors of the combined kernel matrices. Moreover, utilizing the LogDet divergence ensures the positive definiteness of the resulting inferred kernel matrices. Our proposed method of restricting the model covariance matrix via PPCA and PFA resulted to significant improvements in the generalization performance, as shown in our empirical results for the function classification prediction task in yeast proteins.



\section*{Acknowledgment}
This work was supported by JSPS KAKENHI Grant Number
40401236. 


\bibliographystyle{named}
\bibliography{ijcai2018rac-bib}

\tslong{
\appendix

\section{Proofs and Derivations}
\subsection{Derivation of E-Step for FA-MKMC}
\label{ss:estep-famkmc}
The joint densities of the complete data,
say $p_{\text{FA}}(\x_{k},\z_{k}\,|\,\vW^{(t-1)}, \vpsi^{(t-1)})$,  
are the Gaussian distribution
with zero mean and covariance matrix given by  
\begin{tsaligned}
  \begin{bmatrix}
    \vM^{(t)} & \vW^{(t-1)}
    \\
    (\vW^{(t-1)})^\top & \vI
  \end{bmatrix}.  
\end{tsaligned}
Let $\vThet^{(t-1)}\coloneqq(\vW^{(t-1)},\vpsi^{(t-1)})$. 
The expectation in the Q-function~\eqref{eq:Qfun-fa}
at $t$-th iteration is taken under 
the posterior distribution of the complete data
given by
\begin{tsaligned}
  q_{t}(\x_{k},\z_{k}) = q_{t}(\z_{k},\vh_{k}|\vv_{k})q(\vv_{k})
\end{tsaligned}
where
\begin{multline}
  q_{t}(\z_{k},\vh_{k}|\vv_{k})
  \\
  =
  p_{\text{FA}}(\z_{k}\,|\,\vh_{k},\vv_{k}, \vThet^{(t-1)})
  p_{\text{FA}}(\vh_{k}\,|\,\vv_{k}, \vThet^{(t-1)}). 
\end{multline}
From the nature of Gaussian, we have
\begin{multline}
  p_{\text{FA}}(\z_{k}\,|\,\vh_{k},\vv_{k}, \vThet^{(t-1)})
  \\
  =
  \cN\left( \z_{k}\,;\,
            \vB^{(t)}\x_{k},
              \vI-\vB^{(t)}\vW^{(t-1)}\right). 
\end{multline}
In E-step, the terms of the second moments
contained in the Q-function are computed.
Using a similar derivation described in
\cite{RivLemKat17a}, 
the second moment of $\x_{k}$
under $q_{t}(\x_{k},\z_{k})$ is expressed as 
\begin{tsaligned}
  &\bE_{q_{t}(\x_{k},\z_{k})}\left[\x_{k}\x_{k}^\top\right] = \vQ^{(k)}.
\end{tsaligned}
This allows us to write the second moment of $\z_{k}\x_{k}^\top$ as
\begin{tsaligned}
  &\bE_{q_{t}(\x_{k},\z_{k})}\left[\x_{k}\z_{k}^\top\right]
  =
  \bE_{q_{t}(\x_{k})}\left[\x_{k}
    \bE_{q_{t}(\z_{k}\,|\,\x_{k})}\left[ \z_{k}^\top\right]\right]
  \\
  &=
  \bE_{q_{t}(\x_{k})}\left[\x_{k}
    \x_{k}^\top(\vB^{(t)})^\top\right]
  = \vQ^{(k)}(\vB^{(t)})^\top,
\end{tsaligned}
and the second moment of $\z_{k}\z_{k}^\top$ as
\begin{tsaligned}
  &\bE_{q_{t}(\x_{k},\z_{k})}\left[\z_{k}\z_{k}^\top\right]
  =
  \bE_{q_{t}(\x_{k})}\left[
    \bE_{q_{t}(\z_{k}\,|\,\x_{k})}\left[ \z_{k}\z_{k}^\top\right]\right]
  \\
  &=
  \bE_{q_{t}(\x_{k})}\left[
    \vB^{(t)}\x_{k}\x_{k}^\top(\vB^{(t)})^\top\right]
  = \vB^{(t)}\vQ^{(k)}(\vB^{(t)})^\top ,
\end{tsaligned}
where
\begin{tsaligned}
q_{t}(\x_{k})\coloneqq
\int q_{t}(\x_{k},\z_{k})d\z_{k}. 
\end{tsaligned}

\subsection{Derivation of M-Step for FA-MKMC}
\label{ss:mstep-famkmc}
The derivatives of the function $Q^{\text{FA}}_{t}(\vW,\vpsi)$
with respect to $\vW$ and $\psi_{i}^{-1}$
for $i=1,\dots,\ell$ are
written as
\begin{tsaligned}
\frac{\partial \cQ^{\text{FA}}_{t}(\vW,\vpsi)}{\partial \vW}
=
K\diag(\vpsi)^{-1}
\left(
\vS_{xz}^{(t)} - \vW\vS_{zz}^{(t)}
\right)
\end{tsaligned}
and
\begin{multline}
\frac{\partial \cQ^{\text{FA}}_{t}(\vW,\vpsi)}{\partial \psi_{i}^{-1}}
=
\\
\frac{K}{2}
\psi_{i}
-
\frac{K}{2}
\left[
\vS_{xx}^{(t)} - \vS_{xz}^{(t)}\left(\vS_{zz}^{(t)}\right)^{-1}\left(\vS_{xz}^{(t)}\right)^\top
\right]_{i,i}. 
\end{multline}
Setting them to zero yields \eqref{eq:W-psi-upd-famkmc}. 

\subsection{Proof of Proposition~\ref{prop:FAMKMC-monodecreasing}}
\label{ss:proof-prop:FAMKMC-monodecreasing}
Let $\vThet\coloneqq(\vW,\vpsi)$. 
The objective function of the model update step
in FA-MKMC algorithm can be rewritten as
\begin{tsaligned}
&J(\cH^{(t)}\,,\vW,\vpsi)
\\
&=
\sum_{k=1}^{K}\bE_{q_{t}(\x_{k})}\left[ \log q_{t}(\x_{k}) \right]
\\
&\qquad\qquad
-
\sum_{k=1}^{K}\bE_{q_{t}(\x_{k})}
\left[ \log p_{\text{FA}}( \x_{k}\,|\,\vThet ) \right]
\\
&=
\sum_{k=1}^{K}\bE_{q_{t}(\x_{k})}\left[ \log q_{t}(\x_{k}) \right]
\\
&\qquad\qquad
-
\sum_{k=1}^{K}\bE_{q_{t}(\x_{k},\z_{k})}
\left[ \log
\frac{p_{\text{FA}}( \x_{k},\z_{k}\,|\,\vThet )}{p_{\text{FA}}( \z_{k}\,|\,\x_{k},\vThet)} \right],  
\end{tsaligned}
where the last equality follows because
for any $k \in \{1,\dots,K\}$ and any $\z_{k}\in\bR^{q}$,
it holds that 
\begin{tsaligned}
p_{\text{FA}}( \x_{k}\,|\,\vThet )
p_{\text{FA}}( \z_{k}\,|\,\x_{k},\vThet)
=
p_{\text{FA}}( \x_{k},\z_{k}\,|\,\vThet ). 
\end{tsaligned}

Meanwhile, the Q-function in the EM algorithm can be
rearranged as
\begin{tsaligned}
&\cQ^{\text{FA}}_{t}(\vW,\vpsi)
=
\sum_{k=1}^{K}\bE_{q_{t}(\x_{k},\z_{k})}
\left[ \log p_{\text{FA}}( \x_{k},\z_{k}\,|\,\vThet ) \right]
\\
&=
\sum_{k=1}^{K}\bE_{q_{t}(\x_{k},\z_{k})}
\left[ \log
\frac{p_{\text{FA}}( \x_{k},\z_{k}\,|\,\vThet )}{p_{\text{FA}}( \z_{k}\,|\,\x_{k},\vThet)} \right]
\\
&\qquad
-
\sum_{k=1}^{K}\bE_{q_{t}(\x_{k})}
\left[
\bE_{q_{t}(\z_{k}|\x_{k})}
\left[
\log
p_{\text{FA}}( \z_{k}\,|\,\x_{k},\vThet)
\right]\right]
\\
&=
-J(\cH^{(t)}\,,\vW,\vpsi)
+
\sum_{k=1}^{K}\bE_{q_{t}(\x_{k})}\left[ \log q_{t}(\x_{k}) \right]
\\
&\qquad
-
\sum_{k=1}^{K}\bE_{q_{t}(\x_{k})}
\left[
\bE_{q_{t}(\z_{k}|\x_{k})}
\left[
\log
p_{\text{FA}}( \z_{k}\,|\,\x_{k},\vThet)
\right]\right..
\end{tsaligned}

Recall that $(\vW^{(t)},\vpsi^{(t)})$ is determined to be
the value maximizing $\cQ^{\text{FA}}_{t}(\vW,\vpsi)$,
which implies that, 
if we denote by $\text{KL}(\cdot || \cdot)$ the
Kullback-Leibler divergence between two
probabilistic density functions, we have
\begin{tsaligned}
0 &\le
\cQ^{\text{FA}}_{t}(\vW^{(t)},\vpsi^{(t)})
-
\cQ^{\text{FA}}_{t}(\vW^{(t-1)},\vpsi^{(t-1)})
\\
&=
J(\cH^{(t)}\,,\vW^{(t-1)},\vpsi^{(t-1)})
-J(\cH^{(t)}\,,\vW^{(t)},\vpsi^{(t)})
\\
&-
\sum_{k=1}^{K}\bE_{q_{t}(\x_{k})}
\left[
\text{KL}
\left(
q_{t}(\z_{k}|\x_{k}) || p_{\text{FA}}( \z_{k}\,|\,\x_{k},\vThet^{(t)})
\right)
\right]
\\
&\le
J(\cH^{(t)}\,,\vW^{(t-1)},\vpsi^{(t-1)})
-J(\cH^{(t)}\,,\vW^{(t)},\vpsi^{(t)}),
\end{tsaligned}
where the second inequality follows since
Kullback-Leibler divergence is nonnegative.
The proof is now complete.

}

\end{document}